\documentclass{article}

\usepackage{graphicx}
\usepackage{subfigure}

\usepackage[utf8]{inputenc} 
\usepackage[T1]{fontenc}    

\usepackage{hyperref}       
\hypersetup{colorlinks, citecolor=[rgb]{0,0.08,0.65}, urlcolor=[rgb]{0,0.08,0.65}, linkcolor=[rgb]{0,0.08,0.65}, filecolor=[rgb]{0,0.08,0.65}}

\usepackage{url}            
\usepackage{booktabs}       
\usepackage{bm}
\usepackage{dsfont}
\usepackage{dsfont} 
\usepackage{amsfonts}       
\usepackage{nicefrac}       
\usepackage{microtype}      
\usepackage{amsmath}
\usepackage{amsthm}
\usepackage{amssymb}
\usepackage{amsfonts}
\usepackage{algorithm,algorithmic}
\usepackage{multirow}

\usepackage{amsmath,amsfonts,bm}

\def\eqref#1{equation~\ref{#1}}

\def\1{\bm{1}}

\def\vtheta{{\bm{\theta}}}
\def\vbeta{{\bm{\beta}}}

\def\vphi{{\bm{\phi}}}
\def\vpsi{{\bm{\psi}}}

\def\vf{{\bm{f}}}

\def\vh{{\bm{h}}}

\def\vx{{\bm{x}}}
\def\vy{{\bm{y}}}
\def\vz{{\bm{z}}}

\def\mK{{\bm{K}}}

\def\mX{{\bm{X}}}

\def\mZ{{\bm{Z}}}

\def\mLambda{{\bm{\Lambda}}}

\DeclareMathAlphabet{\mathsfit}{\encodingdefault}{\sfdefault}{m}{sl}
\SetMathAlphabet{\mathsfit}{bold}{\encodingdefault}{\sfdefault}{bx}{n}

\newcommand{\KL}{D_{\mathrm{KL}}}

\usepackage[accepted]{icml2019}

\usepackage{enumitem}
\setlist[itemize]{topsep=0pt,itemsep=-1ex,partopsep=1ex,parsep=1ex,leftmargin=15pt}
\setlist[enumerate]{topsep=0pt,itemsep=-1ex,partopsep=1ex,parsep=1ex,leftmargin=15pt}

\newlength{\figwidthone}
\newlength{\figwidththree}
\icmltitlerunning{Hybrid Models with Deep and Invertible Features}

\def\indicator{\mathds{1}}

\begin{document}
\cfoot{\thepage}
\setlength{\footskip}{3em}

\twocolumn[
\icmltitle{Hybrid Models with Deep and Invertible Features}



\icmlsetsymbol{equal}{*}

\begin{icmlauthorlist}
 \icmlauthor{Eric Nalisnick}{equal,dm}
\icmlauthor{Akihiro Matsukawa}{equal,dm}
\icmlauthor{Yee Whye Teh}{dm}
\icmlauthor{Dilan Gorur}{dm}
\icmlauthor{Balaji Lakshminarayanan}{dm}
\end{icmlauthorlist}

\icmlaffiliation{dm}{DeepMind}
\icmlcorrespondingauthor{Balaji Lakshminarayanan}{balajiln@google.com}

\icmlkeywords{hybrid models, invertible models, uncertainty, selective classification
}

\vskip 0.3in
]



\printAffiliationsAndNotice{\icmlEqualContribution} 

\begin{abstract}

We propose a neural hybrid model consisting of a linear model defined on a set of features computed by a deep, invertible transformation (i.e.\ a normalizing flow). An attractive property of our model is that both $p(\text{\texttt{features}})$,  the density of the features,  and  $p(\text{\texttt{targets}} | \text{\texttt{features}})$, the predictive distribution, can be computed exactly in a single feed-forward pass.   We show that our hybrid model, despite the invertibility constraints, achieves similar accuracy to purely predictive models.  Moreover the generative component remains a good model of the input features despite the hybrid optimization objective.  This offers additional capabilities such as detection of out-of-distribution inputs and enabling semi-supervised learning.  The availability of the exact joint density $p(\text{\texttt{targets}}, \text{\texttt{features}})$ also allows us to compute many quantities readily, making our hybrid model a useful building block for downstream applications of probabilistic deep learning.


\end{abstract}

\section{Introduction}
In the majority of applications, deep neural networks model conditional distributions of the form $p(y|\vx)$, where $y$ denotes a label and $\vx$ features or covariates. However, modeling just the conditional distribution is insufficient in many cases.  For instance, if we believe that the model may be subjected to inputs unlike those of the training data, a model for $p(\vx)$ can possibly detect an outlier before it is passed to the conditional model for prediction.  Thus modeling the joint distribution $p(y, \vx)$ provides a richer and more useful representation of the data.  Models defined by combining a predictive model $p(y | \vx)$ with a generative one $p(\vx)$ are known as \textit{hybrid models} \citep{jaakkola1999exploiting, raina2004classification, lasserre2006principled, kingma2014semi}.  Hybrid models have been shown to be useful for novelty detection \citep{bishop1994novelty},
semi-supervised learning \citep{druck2007semi}, and information regularization \citep{szummer2003information}.


Crafting a hybrid model usually requires training two models, one for $p(y | \vx)$ and one for $p(\vx)$, that share a subset \citep{raina2004classification} or possibly all \citep{mccallum2006multi} of their parameters.  Unfortunately, training a high-fidelity $p(\vx)$ model alone is difficult, especially in high dimensions, and good performance requires using a large neural network \citep{brock2018large}.  Yet principled probabilistic inference is hard to implement with neural networks since they do not admit closed-form solutions and running Markov chain Monte Carlo takes prohibitively long.  Variational inference then remains as the final alternative, and this now introduces a \emph{third} model, which usually serves as the posterior approximation and/or inference network \citep{vae, kingma2014semi}.  To make matters worse, the $p(y | \vx)$ model may require a separate approximate inference scheme, leading to additional computation and parameters.  

In this paper, we propose a neural hybrid model that overcomes many of the aforementioned computational challenges.  Most crucially, our model supports \emph{exact} inference and evaluation of $p(\vx)$.  Furthermore, in the case of regression, Bayesian inference for $p(y| \vx)$ is exact and available in closed-form as well.  
Our model is made possible by leveraging recent advances in deep invertible generative models \citep{rezende2015variational, dinh2016density, kingma2018glow}.  These models are defined by composing invertible functions, and therefore the change-of-variables formula can be used to compute exact densities.  These invertible models have been shown to be expressive enough to perform well on prediction tasks \citep{gomez2017reversible, jacobsen2018revnet}.  We use the invertible function as a natural feature extractor and define a linear model at the level of the latent representation, which is memory-efficient as the bulk of the parameters are shared between $p(\vx)$ and $p(y | \vx)$.  Furthermore, with just \emph{one feed-forward pass} we can obtain both $p(\vx)$ and $p(y | \vx)$, with the only additional cost being the log-determinant-Jacobian term required by the change of variables.  While this term could be expensive to compute for general functions, much recent work has been done on defining expressive invertible neural networks with easy-to-evaluate volume elements \citep{dinh2014nice, dinh2016density, kingma2018glow, grathwohl2018ffjord}.   
\newpage
In summary, our contributions are: 
\begin{enumerate}[label={\textbullet}]
    \item Defining a neural hybrid model with exact inference and evaluation of $p(y, \vx)$, which can be computed in one feed-forward pass and without any Monte Carlo approximations.
    \item Evaluating the model's predictive accuracy and uncertainty on both classification and regression problems.
    \item Using the model's natural `reject' rule based on the generative component $p(\vx)$ to filter out-of-distribution (OOD) inputs.
    \item Showing that our hybrid model performs well at semi-supervised classification.
\end{enumerate}

\section{Background}
We begin by establishing notation and reviewing the necessary background material.  We denote matrices with upper-case and bold letters (e.g. $\mX$), vectors with lower-case and bold (e.g. $\vx$), and scalars with lower-case and no bolding (e.g. $x$).  Let the collection of all observations be denoted $\mathcal{D} = \{ \mX, \vy\} = \{ (\vx_{n}, y_{n})_{n=1}^{N}\} $ with $\vx$ representing a vector containing features and $y$ a scalar representing the corresponding label.  We define a predictive model's density function to be $p(y | \vx; \vtheta)$ and a generative density to be $p(\vx; \vtheta)$, where $\vtheta \in \boldsymbol{\Theta}$ are the shared model parameters.  Let the joint likelihood be denoted $p(\vy, \mX;\vtheta) = \prod_{n=1}^{N} p(y_{n} | \vx_{n}; \vtheta) p(\vx_{n}; \vtheta)$. 

\subsection{Invertible Generative Models} 
Deep invertible transformations are the first key building block in our approach.  These are simply high-capacity, bijective transformations with a tractable Jacobian matrix and inverse.  The best known models of this class are the \textit{real non-volume preserving} (RNVP) transform \citep{dinh2016density} and its recent extension, the \textit{Glow} transform \citep{kingma2018glow}.  The bijective nature of these transforms is crucial as it allows us to employ the change-of-variables formula for \emph{exact} density evaluation: \begin{align}\label{eq:cov}
 \log p_{x}(\vx) = \log p_{z}(f(\vx; \vphi)) + \log \left| \frac{\partial \vf_{\vphi}}{ \partial \vx} \right|
\end{align} where $f(\cdot; \vphi)$ denotes the transform with parameters $\vphi$, $\left| \partial \vf / \partial \vx \right|$ the determinant of the Jacobian of the transform, and $p_{z}(\vz=f(\cdot; \vphi))$ a distribution on the latent variables computed from the transform.  The modeler is free to choose $p_{z}$, and therefore it is often set as a factorized standard Gaussian for computational simplicity.  The \textit{affine coupling layer} (ACL) \citep{dinh2016density} is the key building block used by RNVP and Glow to define $f(\cdot; \vphi)$.  It consists of transforming half of the representation with translation and scaling operations and copying the other half forward to the output.  See 
Appendix~\ref{sec:acl} 
in the supplementary material for a detailed description of the ACL. 
Glow \citep{kingma2018glow} introduces $1 \times 1$ convolutions between ACLs.  The parameters $\vphi$ are estimated via maximizing the exact log-likelihood $\log p(\mX; \vphi)$.  

While the invertibility requirements imposed on $f$ may seem too restrictive to define an expressive model, recent work using invertible transformations for classification \citep{jacobsen2018revnet} reports metrics comparable to non-invertible residual networks, even on challenging benchmarks such as ImageNet, and recent work by \citet{kingma2018glow} has shown that invertible generative models can produce sharp samples.  Sampling from a flow is done by first sampling from the latent distribution and then passing that sample through the inverse transform: $\hat{\vz} \sim p_{z},$ $\hat{\vx} = f^{-1}(\hat{\vz})$.

\subsection{Generalized Linear Models}
\textit{Generalized linear models} (GLMs) \citep{nelder1972generalized} are the second key building block that we employ.  They model the expected response $y$ as follows: \begin{align}
    \mathbb{E}[y_{n} | \vz_{n}] = g^{-1}\left( \boldsymbol{\beta}^{T} \vz_{n} \right)
\end{align} where $\mathbb{E}[y | \vz]$ denotes the expected value of $y_{n}$, $\vbeta$ a $\mathbb{R}^{d}$ vector of parameters, $\vz_{n}$ the covariates, and $g^{-1}(\cdot)$ a \textit{link function} such that $g^{-1}: \mathbb{R} \mapsto \mu_{y | \vz}$.  For notational convenience, we assume a scalar bias $\beta_{0}$ has been subsumed into $\vbeta$.  A Bayesian GLM could be defined by specifying a prior $p(\vbeta)$ and computing the posterior $p(\vbeta|\vy, \mZ)$.  When the link function is the identity (i.e. simple linear regression) and $\vbeta \sim \text{N}(\mathbf{0}, \mLambda^{-1})$, then the posterior distribution is available in closed-form: \begin{align}
    p(\vbeta | \vy, \mX) = \text{N}\left( \frac{\mX^{T}\vy}{\mX^{T}\mX +  \sigma_{0}^{2}\mLambda}, \  \frac{\sigma_{0}^{2}}{\mX^{T}\mX +  \sigma_{0}^{2} \mLambda} \right)
\end{align} where $\sigma_{0}$ is the response noise.  In the case of logistic regression, the posterior is no longer conjugate but can be closely approximated \citep{jaakkola1997variational}.

\section{Combining Deep Invertible Transforms and Generalized Linear Models}

We propose a neural hybrid model consisting of a deep invertible transform coupled with a GLM.  Together the two define a \emph{deep} predictive model with both the ability to compute $p(\vx)$ and $p(y | \vx)$ \emph{exactly}, in a \emph{single feed-forward pass}.  The model defines the following joint distribution over a label-feature pair $(y_{n}, \vx_{n})$: \begin{align}\begin{split}
    & p(y_{n},  \vx_{n}; \vtheta) = p(y_{n}| \vx_{n}; \vbeta, \vphi) \ \  p(\vx_{n};  \vphi) \\ &= p(y_{n} | f(\vx_{n};\vphi); \vbeta) \ \  p_{z}(f(\vx_{n};  \vphi)) \ \  \left| \frac{\partial \vf_{\vphi}}{ \partial \vx_{n}} \right|
\end{split}\end{align} where $\vz_{n} = f(\vx_{n}, \vphi)$ is the output of the invertible transformation, $p_{z}(\vz)$ is the latent distribution (also referred to as the prior or base distribution), and $p(\mathbf{y}_{n} | f(\vx_{n};\vphi); \vbeta)$ is a GLM with the latent variables serving as its input features.  For simplicity, we assume a factorized latent distribution $p(\vz)=\prod_d p(z_d)$, following previous work \citep{dinh2016density, kingma2018glow}.  Note that $\vphi=\{\vphi_{t,l}, \vphi_{s,l} \}_{l=1}^{L}$ are the parameters of the generative model and that $\vtheta=\{\vphi, \vbeta\}$ are the parameters of the joint model.  Sharing $\vphi$ between both components allows the conditional distribution to influence the generative distribution and vice versa.  
 We term the proposed neural hybrid model the \emph{deep invertible generalized linear model} (DIGLM).  Given labeled training data $\{(\vx_n,y_n)\}_{n=1}^N$ sampled from the true distribution of interest $p^*(\vx,y)$, the DIGLM can be trained by maximizing the exact joint log-likelihood, i.e.\ 
\begin{align*} 
    \mathcal{J}(\vtheta) = \log p(\vy, \mX; \vtheta) = \sum_{n=1}^{N} \log p(y_{n}, \vx_{n}; \vtheta),
\end{align*} 
via gradient ascent.  As per the theory of maximum likelihood, maximizing this log probability is equivalent to minimizing the Kullback-Leibler (KL) divergence between the true joint distribution and the model: $\KL\bigl(p^*(\vx,y)\| p_{\vtheta}(\vx,y)\bigr)$.  

\begin{figure}
\centering
\includegraphics[width=0.65\columnwidth]{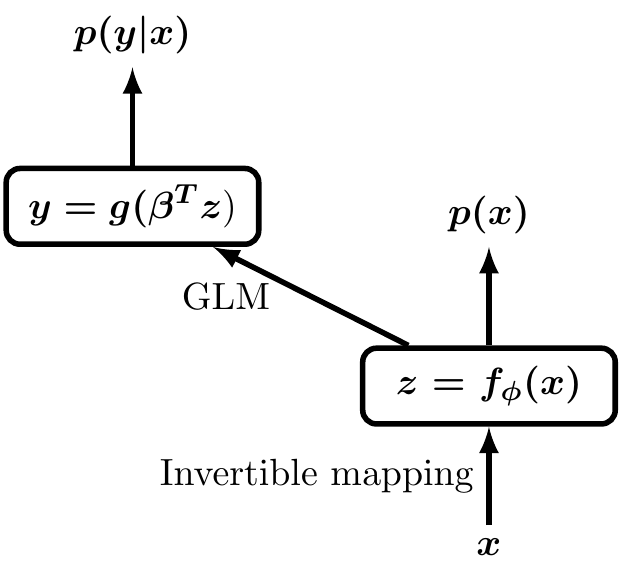}
\caption{\textit{Model Architecture}.  The diagram above shows the DIGLM's computational pipeline, which is comprised of a GLM stacked on top of an invertible generative model. The model parameters are $\vtheta=\{\vphi, \vbeta\}$ of which $\vphi$ is shared between the generative and predictive model, and $\vbeta$ denotes parametrizes the GLM in the predictive model. 
}
\label{wrap-fig:1}
\end{figure}

Figure \ref{wrap-fig:1} shows a diagram of the DIGLM.  We see that the computation pipeline is essentially that of a traditional neural network but one defined by stacking ACLs.  The input $\vx$ first passes through $f_{\vphi}$, and the latent representation and the stored Jacobian terms are enough to compute $p(\vx)$.  In particular, evaluating $p_{z}(f(\vx_{n};  \vphi))$ has an $\mathcal{O}(D)$ run-time cost for factorized distributions, and $\left| \partial \vf_{\vphi} /  \partial \vx_{n} \right|$ has a $\mathcal{O}(LD)$ run-time for RNVP architectures, where $L$ is the number of affine coupling layers and $D$ is the input dimensionality.  Evaluating the predictive model adds another $\mathcal{O}(D)$ cost in computation, but this cost will be dominated by the prerequisite evaluation of $f_{\vphi}$. 

\paragraph{Weighted Objective}
In practice we found the DIGLM's performance improved by introducing a scaling factor on the contribution of $p(\vx)$.  The factor helps control for the effect of the drastically different dimensionalities of $y$ and $\vx$.  We denote this modified objective as: \begin{align}\begin{split}
    \mathcal{J}_{\lambda}(\vtheta) &= \sum_{n=1}^{N} \Bigl(\log p(y_{n} | \vx_{n}; \vbeta, \vphi) + \lambda \log p( \vx_{n};  \vphi)\Bigr)
\end{split}\end{align} where $\lambda$ is the scaling constant.  
Weighted losses are commonly used in hybrid models \citep{lasserre2006principled, mccallum2006multi, kingma2014semi, tulyakov2017hybrid}.  Yet in our particular case, we can interpret the down-weighting as encouraging robustness to input variations.  Down-weighting the contribution of $\log p( \vx_{n};  \vphi)$ can be considered a Jacobian-based regularization penalty.  To see this, notice that the joint likelihood rewards maximization of $\left| \partial \vf_{\vphi} /  \partial \vx_{n} \right|$, thereby encouraging the model to increase the $\partial f_{d} / \partial x_{d}$ derivatives (i.e. the diagonal terms).  This optimization objective stands in direct contrast to a long history of gradient-based regularization penalties \citep{girosi1995regularization, bishop1995training, rifai2011contractive}, which  add the Frobenius norm of the Jacobian as a \emph{penalty} to a loss function (or negative log-likelihood).  Thus, we can interpret the de-weighting of $\left| \partial \vf_{\vphi} /  \partial \vx_{n} \right|$ as adding a Jacobian regularizer with weight $\tilde{\lambda} = (1-\lambda)$.  If the latent distribution term is, say, a factorized Gaussian, the variance can be scaled by a factor of $1/\lambda$ to introduce regularization only to the Jacobian term. 

\subsection{Semi-supervised learning}\label{sec:ss}
As mentioned in the introduction, having a  representation of the joint density enables the model to be trained on data sets that do not have a label for every feature vector---i.e.\ semi-supervised data sets.  When a label is not present, the principled approach is to integrate out the variable: 
\begin{align}\label{eq:semi} 
    \int_{y} p(y, \vx; \vtheta, \vphi) \ dy =  p(\vx; \vphi) \int_{y} p(y | \vx; \vtheta) \ dy =  p(\vx; \vphi).
\end{align} 
Thus we should use the unpaired $\vx$ observations to train just the generative component.  

\subsection{Selective Classification}\label{sec:selective:classification}

Equation \ref{eq:semi} 
above also suggests a strategy for evaluating the model in real-world situations.  One can imagine the DIGLM being deployed as part of a user-facing system and that we wish to have the model `reject' inputs that are unlike the training data.  In other words, the inputs are anomalous with respect to the training distribution, and we cannot expect the $p(y|\vx)$ component to make accurate predictions when $\vx$ is not drawn from the training distribution.  In this setting we have access only to the user-provided features $\vx^{*}$, and thus should evaluate by way of Equation \ref{eq:semi} again, computing $p(\vx^{*}; \vphi)$.  This observation then leads to the natural rejection rule:
\begin{align}\label{eq:rej}
    \text{if } \ p(\vx^{*}; \vphi) < \tau, \text{ then } \  \text{\texttt{reject} }  \vx^{*} 
\end{align} where $\tau$ is some threshold, which we propose setting as $\tau = \min_{\vx \in \mathcal{D}} p(\vx; \vphi) - c$ where the minimum is taken over the training set and $c$ is a free parameter providing slack in the margin.  
When rejecting a sample, we output the unconditional $p(y)$, e.g. uniform probabilities for classification problems, hence the prediction for $\vx^{*}$ is given by 
\begin{align}\label{eq:safeprediction}
    p(y) \ \indicator[p(\vx^{*}; \vphi)<\tau] + p(y|\vx^{*}) \ \indicator[p(\vx^{*}; \vphi) \geq\tau]
\end{align} where $\indicator[\cdot]$ denotes an indicator function.
Similar generative-model-based rejection rules have been proposed previously \citep{bishop1994novelty}. This idea is also known as \emph{selective classification} or \emph{classification with a reject option} \citep{hellman1970nearest, cordella1995method, fumera2002support, herbei2006classification, geifman2017selective}.  

\section{Bayesian Treatment}
We next describe a Bayesian treatment of the DIGLM, deriving some closed-form quantities of interest and discussing connections to Gaussian processes.  The Bayesian DIGLM (B-DIGLM) is defined as follows: \begin{align*}f(\vx; \vphi) \sim p(\vz), \ \ \vbeta \sim p(\vbeta),  \ \ y_{n} \sim p(y_{n}| f(\vx_{n}; \vphi), \boldsymbol{\beta}).
\end{align*} The material difference from the earlier formulation is that a prior $p(\vbeta)$ is now placed on the regression parameters.  The B-DIGLM defines the joint distribution of three variables---$p(y_{n},  \vx_{n}, \vbeta ; \vphi)$---and to perform proper Bayesian inference, we should marginalize over $p(\vbeta)$ when training, resulting in the modified objective: \begin{align}\label{marg_joint}\begin{split}
  & p(y_{n},  \vx_{n}; \vphi) =  \int_{\vbeta} p(y_{n},  \vx_{n}, \vbeta ; \vphi) \ d \vbeta  \\ & = \int_{\vbeta} p(y_{n}| \vx_{n}; \vphi, \vbeta) p(\vbeta) \ d \vbeta \ \  p(\vx_{n};  \vphi) \\ &= p(y_{n} | f(\vx_{n};\vphi)) \ \  p_{z}(f(\vx_{n};  \vphi)) \ \  \left| \frac{\partial \vf_{\vphi}}{ \partial \vx_{n}} \right|
\end{split}\end{align} where $p(y_{n} | f(\vx_{n};\vphi))$ is the marginal likelihood of the regression model.  

While $p(y_{n} | f(\vx_{n};\vphi))$ is not always available in closed-form, it is in 
some 
cases.  For instance, if we assume that the likelihood model is Gaussian as in linear regression, and that $\vbeta$ is given a zero-mean Gaussian prior, i.e. $$ p(y_{n} | \vz_{n}, \boldsymbol{\beta}) = \text{N}(y_{n}; \boldsymbol{\beta}^{T}\vz_{n}, \sigma_{0}^{2}), \ \ \ \boldsymbol{\beta} \sim \text{N}(\mathbf{0}, \lambda^{-1}\mathbb{I}) $$ then the marginal likelihood can be written as: 
\begin{align}\label{lr_opt_obj} 
&\log p(y_{n} | f(\vx_{n};\vphi)) \nonumber\\ &= \log \text{N}\left( \vy; \ \mathbf{0}, \   \sigma_{0}^{2}\mathbb{I} + \lambda^{-1} \mZ_{\vphi}\mZ_{\vphi}^{T} \right) \\&\propto -\vy^{T} (\sigma_{0}^{2}\mathbb{I} + \lambda^{-1} \mZ_{\vphi}\mZ_{\vphi}^{T})^{-1} \vy - \log \left| \sigma_{0}^{2}\mathbb{I} + \lambda^{-1} \mZ_{\vphi}\mZ_{\vphi}^{T}\right| \nonumber
\end{align}
where $\mZ_{\vphi}$ is the matrix of all latent representations, which we subscript with $\vphi$ to emphasize that it depends on the invertible transform's parameters.  

\paragraph{Connection to Gaussian Processes}  
From Equation \ref{lr_opt_obj} we see that B-DIGLMs are related to \textit{Gaussian processes} (GPs) \citep{rasmussen2006gaussian}.  GPs are defined through their kernel function $k(\vx_{i}, \vx_{j}; \vpsi)$, which in turn characterizes the class of functions represented.  The marginal likelihood under a GP is defined as \begin{align}\label{lr_opt_obj2} 
\log p(y_{n} | \vx_{n}; \vpsi) \nonumber \propto -\vy^{T} (\sigma_{0}^{2}\mathbb{I} + \mK_{\vpsi})^{-1} \vy - \log \left| \sigma_{0}^{2}\mathbb{I} + \mK_{\vpsi}\right| \nonumber
\end{align} with $\vpsi$ denoting the kernel parameters.  Comparing this equation to the B-DIGLM's marginal likelihood in Equation \ref{lr_opt_obj}, we see that they become equal by setting $\mK_{\vpsi} = \lambda^{-1} \mZ_{\vphi}\mZ_{\vphi}^{T}$, and thus we have the implied kernel $k(\vx_{i}, \vx_{j}) = \lambda^{-1} f(\vx_{i};\vphi)^{T} f(\vx_{j};\vphi)$.  Perhaps there are even deeper connections to be made via \textit{Fisher kernels} \citep{jaakkola1999exploiting} or \textit{probability product kernels} \citep{jebara2004probability}---kernel functions derived from generative models---but we leave this investigation to future work.

\paragraph{Approximate Inference} 
If the marginal likelihood is not available in closed form, then we must resort to approximate inference.  In this case, understandably, our model loses the ability to compute exact marginal likelihoods.  We can use one of the many lower bounds developed for variational inference to bypass the intractability.  Using the usual variational Bayes evidence lower bound (ELBO) \citep{jordan1999introduction}, we have \begin{align}\begin{split} 
\log p(y_{n} | f(\vx_{n};\vphi)) \ge \mathbb{E}_{q(\vbeta)}&\left[ p(y_{n}| \vx_{n}; \vphi, \vbeta) \right] \\ & - \text{KLD}\left[q(\vbeta) || p(\vbeta) \right]
\end{split}\end{align} where $q(\vbeta)$ is a variational approximation to the true posterior.  We leave  thorough investigation of approximate inference to future work, and in the experiments we use either conjugate Bayesian inference or point estimates for $\vbeta$.   

One may ask: why stop the Bayesian treatment at the predictive component?  Why not include a prior on the flow's parameters as well?  This could be done, but \citet{riquelme2018deep} showed that Bayesian linear regression with deep features (i.e. computed by a deterministic neural network) is highly effective for contextual bandit problems, which suggests that capturing the uncertainty in prediction parameters $\vbeta$ is more important than the uncertainty in the representation parameters $\vphi$.

\begin{figure*}[htbp]
\subfigure[Gaussian Process]{  \includegraphics[width=\figwidththree]{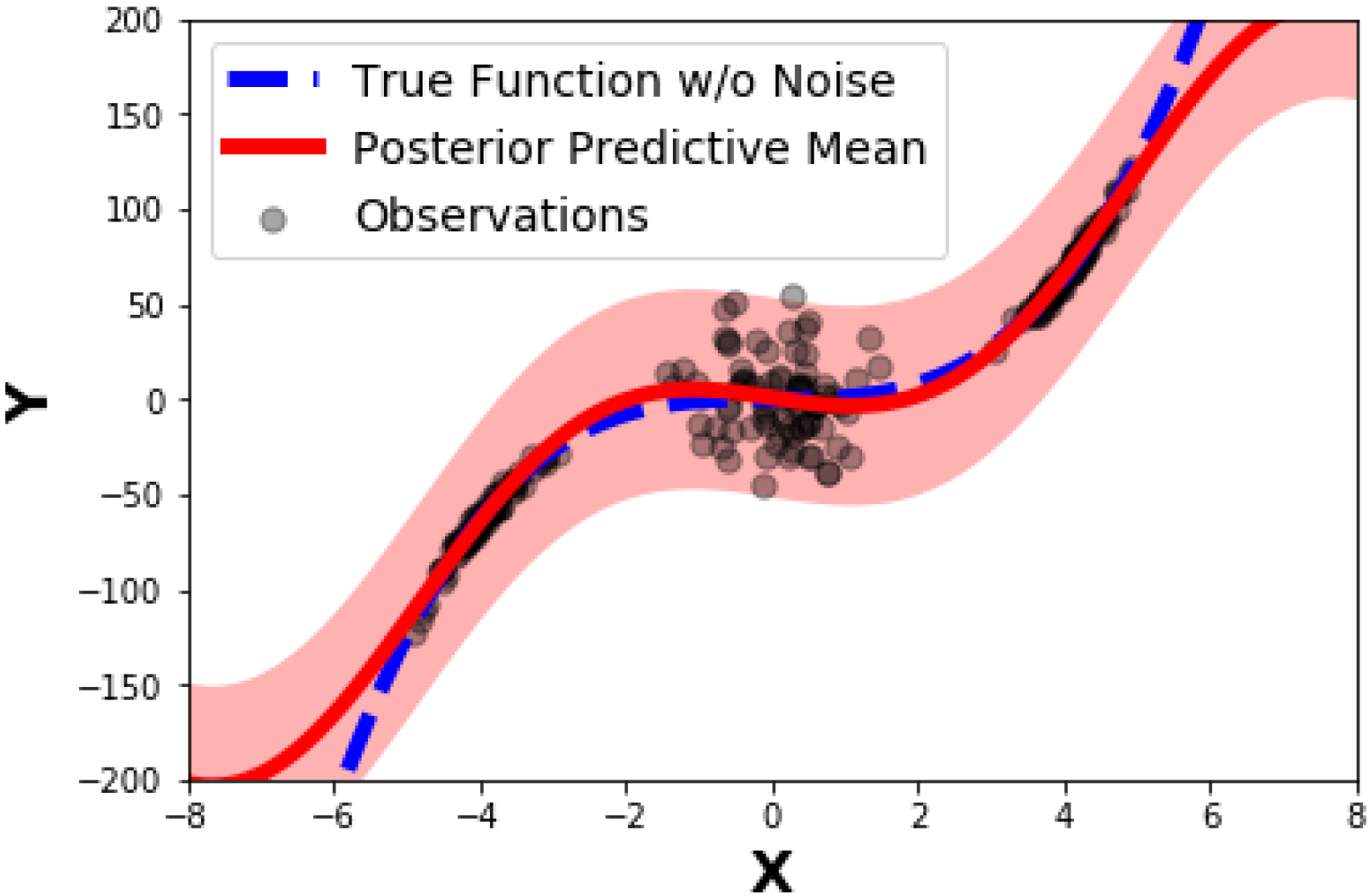}\label{fig:1dgp}}
\hfill
\subfigure[B-DIGLM $\ p(y|\vx)$]{ \includegraphics[width=\figwidththree]{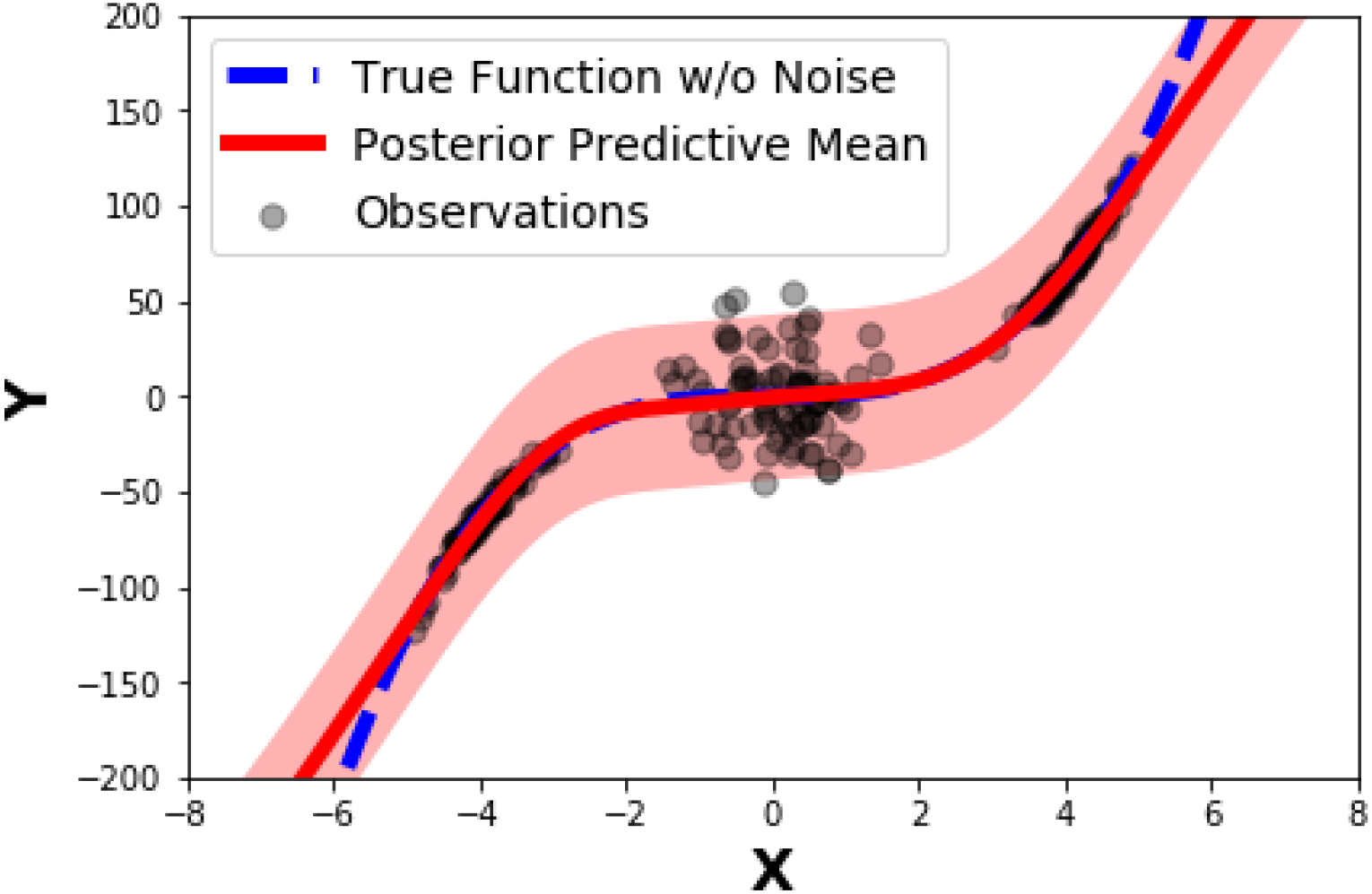}\label{fig:1ddig}}
\hfill
\subfigure[B-DIGLM $\ p(\vx)$]{ \includegraphics[width=\figwidththree]{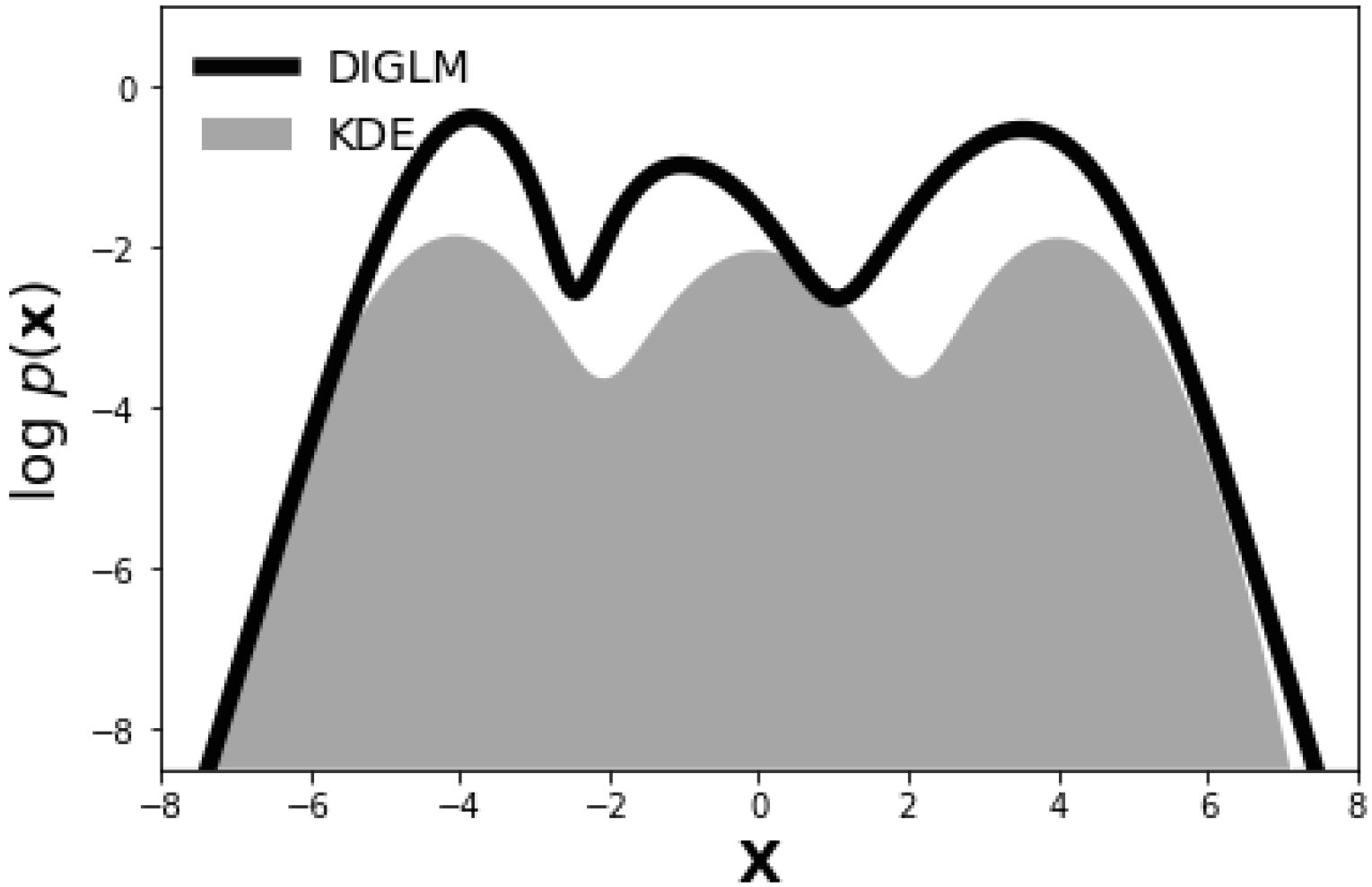}\label{fig:1dpx}}
 \caption{\textit{1-dimensional Regression Task}.  We construct a toy regression task by sampling $x$-observations from a Gaussian mixture model and then assigning responses $y = x^{3} + \epsilon$ with $\epsilon$ being  heteroscedastic noise.  Subfigure (a) shows the function learned by a Gaussian process and (b) shows the function learned by the Bayesian DIGLM.  Subfigure (c) shows the $p(x)$ density learned by the same DIGLM (black line) and compares it to a KDE (gray shading).}
\label{toy_regression_results}
\end{figure*}

\section{Related Work}
We are unaware of any work that uses normalizing flows as the generative component of a hybrid model.  The most related work is the class conditional variant of Glow \citep[Appendix D]{kingma2018glow}.  For this model, \citet{kingma2018glow} use class-conditional latent distributions and introduce a (down-weighted) classification loss to the penultimate layer of the flow.  However, they do not evaluate the model for its predictive capabilities and instead (qualitatively) evaluate its class-conditional generative abilities.  

While several works have studied the trade-offs between generative and predictive models \citep{efron1975efficiency, ng2002discriminative}, \citet{jaakkola1999exploiting} were perhaps the first to meaningfully combine the two, using a generative model to define a kernel function that could then be employed by classifiers such as SVMs.  \citet{raina2004classification} took the idea a step further, training a subset of a naive Bayes model's parameters with an additional predictive objective.  \citet{mccallum2006multi} extended this framework to train all parameters with both generative and predictive objectives.  \citet{lasserre2006principled} showed that a simple convex combination of the generative and predictive objectives does not necessarily represent a unified model and proposed an alternative prior that better couples the parameters.  \citet{druck2007semi} empirically compared \citet{lasserre2006principled}'s and \citet{mccallum2006multi}'s hybrid objectives specifically for semi-supervised learning.  Recent advances in deep generative models and stochastic variational inference have allowed the aforementioned frameworks to include neural networks as the predictive and/or generative components.  Deep neural hybrid models haven been defined by (at least) \citet{kingma2014semi}, \citet{maaloe2016auxiliary}, \citet{kuleshov2017deep}, \citet{tulyakov2017hybrid}, 
and \citet{gordon2017bayesian}.  However, these models, unlike ours, require approximate inference to obtain the $p(\vx)$ component.  

As mentioned in the introduction, invertible residual networks have been shown to perform as well as non-invertible architectures on popular image benchmarks \citep{gomez2017reversible, jacobsen2018revnet}.  While the change-of-variables formula could be calculated for these models, it is computationally difficult to do so, which prevents their application to generative modeling.  The concurrent work of \citet{behrmann2018invertible} shows how to preserve invertibility in general residual architectures and describes a stochastic approximation of the volume element to allow for high-dimensional generative modeling.  Hence their work could be used to define a hybrid model similar to ours, which they mention as area for future work.


\section{Experiments}  We now report experimental findings for a range of regression and classification tasks.  Unless otherwise stated, we used the Glow architecture \citep{kingma2018glow} to define the DIGLM's invertible transform and factorized standard Gaussian distributions as the latent prior $p(\vz)$.  

\subsection{Regression on Simulated Data}
We first report a one-dimensional regression task to provide an intuitive demonstration of the DIGLM.  We draw $x$-observations from a Gaussian mixture with parameters $\mu=\{ -4, 0, +4\}$, $\sigma=\{.4, .6, .4 \}$, and equal component weights.  We simulate responses with the function $y = x^{3} + \epsilon(k)$ where $\epsilon(k)$ denotes observation noise as a function of the mixture component $k$.  Specifically we chose $\epsilon(k) \sim \mathds{1}[k \in \{1,3\}]N(0, 3) + \mathds{1}[k=2]N(0, 20)$.  We train a B-DIGLM on 250 observations sampled in this way, use standard Normal priors for $p(\vz)$ and $p(\vbeta)$, and three planar flows \citep{rezende2015variational} to define $f(\vx)$.  We compare this model to a Gaussian process (GP) and a kernel density estimate (KDE), which both use squared exponential kernels.  

Figure~\ref{fig:1dgp} shows the predictive distribution learned by the GP, and Figure~\ref{fig:1ddig} shows the DIGLM's predictive distribution.  We see that the models produce similar results, with the only conspicuous difference being the GP has a stronger tendency to revert to its mean at the plot's edges. Figure~\ref{fig:1dpx} shows the $p(x)$ density learned by the DIGLM's flow component (black line), and we plot it against the KDE (gray shading) for comparison.  The single B-DIGLM is able to achieve comparable results to the separate GP and KDE models.

Thinking back to the rejection rule defined in Equation \ref{eq:rej}, this result, albeit on a toy example, suggests that density thresholding would work well in this case.  All data observations fall within $x \in [-5, 6]$, and we see from Figure \ref{fig:1dpx} that the DIGLM's generative model smoothly decays to the left and right of this range, meaning that there does not exist an $x^{*}$ that lies outside the training support and has $ p(\vx^{*}) \ge \min_{\vx \in \mathcal{D}} p(\vx)$.

\subsection{Regression on Flight Delay Data Set}
Next we evaluate the model on a large-scale regression task using the \emph{flight delay} data set \citep{svigp}.  The goal is to predict how long flights are delayed based on eight attributes.  
  Following \citet{dgp}, we train using the first $5$ million data points and use the following $100,000$ as test data. We picked this split not only to illustrate the  scalability of our method, but also due to the fact that the test distribution is known to be slightly different from training, which poses challenges of non-stationarity. We evaluate the performance by measuring the root mean squared error (RMSE) and the negative log-likelihood (NLL).
  
\begin{figure}[h]
\centering
\includegraphics[width=\figwidthone]{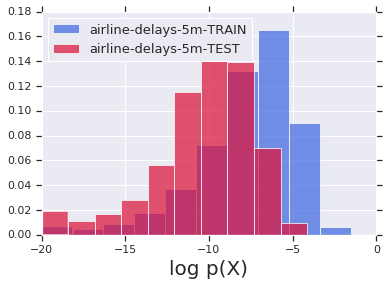}
\caption{Histogram of $\log p(\vx)$ on the 
flight delay data set.  The leftward shift in the test set (red) shows that our DIGLM model is able to detect  covariate shift.
}
 \label{regression_results}
\end{figure}

 One could model heteroscedasticity in GLMs using random effects (see 
 Appendix~\ref{sec:extensions} 
 for a discussion), however as a simpler alternative, we follow the solution proposed by \citet{deepensembles} for heteroscedastic regression and set $p(y|\vz)$ to be a two-headed model that predicts both the mean and variance.  We use a RNVP transform as the invertible function where the RNVP blocks use 1-layer network with 100 hidden units, and train using Adam optimizer for 10 epochs with learning rate $10^{-3}$ and batch size $100$. 
To the best of our knowledge, the state-of-the-art (SOTA) performance on this data set is a test RMSE of $38.38$ and a test NLL of $6.91$ \citep{MFR}.
Our hybrid model 
achieves a slightly worse test RMSE of 40.46 but achieves a markedly \textbf{better test NLL of 5.07}.  We believe that this superior NLL stems from the hybrid model's ability to detect the non-stationarity of the data.  Figure~\ref{regression_results} shows a histogram of the $\log p(\vx)$ evaluations for the training data (blue bars) and test data (red bars).  The leftward shift in the red bars confirms that the test data points indeed have lower density under the flow than the training points.    



\subsection{MNIST Classification}


Moving on to classification, we train a DIGLM on MNIST using 16 Glow blocks ($1\times1$ convolution followed by a stack of ACLs) to define the invertible function. Inside of each ACL, we use a 3-layer Highway network \citep{srivastava2015training} with $200$ hidden units to define the translation $t(\cdot; \vphi_{s})$ and scaling $s(\cdot; \vphi_{s})$ operations. We use batch normalization in the networks for simplicity in distributed coordination rather than actnorm as was used by \citet{kingma2018glow}. We use dropout \citep{srivastava2014dropout} before passing $\vz$ to the GLM, and tune dropout rate 
on the validation set. Optimization was done via Adam \citep{kingma2014adam} with a $10^{-4}$ initial learning rate for $100$k steps, then decayed by half at iterations $800$k and $900$k.

We compare the DIGLM to its discriminative component, which is obtained by setting the generative weight to zero (i.e.\ $\lambda=0$). We report test classification error, NLL, and entropy of the predictive distribution. Following \citet{deepensembles}, we evaluate on both the MNIST test set and the NotMNIST test set, using the latter as an out-of-distribution (OOD) set.  The OOD test is a proxy for testing if the model would be robust to anomalous inputs when deployed in a user-facing system.  The results are shown in Table~\ref{tab:table:mnist}.
Looking at the MNIST results, the discriminative model achieves slightly lower test error, but the hybrid model achieves better NLL and entropy. As expected,  $\lambda$ controls the generative-discriminative trade-off with lower values favoring discriminative performance and higher values favoring generative performance.

\begin{table}[htbp]
  \centering
    \resizebox{0.5\textwidth}{!}{
\begin{tabular}{c|c|c|c|c|c|c}
\multirow{2}{*}{Model}
&  \multicolumn{3}{ c }{\textbf{MNIST}} &  \multicolumn{3}{ |c }{\textbf{NotMNIST}} \\ 
 &    BPD $\downarrow$ & error $\downarrow$ & NLL $\downarrow$  &    BPD $\uparrow$  & NLL $\downarrow$ & Entropy $\uparrow$ \\
\midrule
Discriminative ($\lambda=0$) &  81.80* & \textbf{0.67\%}  & 0.082 & 87.74* & 29.27 & 0.130 \\
Hybrid ($\lambda=0.01/D$) & 1.83 & 0.73\% & \textbf{0.035}  & 5.84 & 2.36 & \textbf{2.300} \\
Hybrid ($\lambda=1.0/D$) & 1.26 & 2.22\% & 0.081  & 6.13 & \textbf{2.30} & \textbf{2.300} \\
Hybrid ($\lambda=10.0/D$) & \textbf{1.25} & 4.01\% & 0.145  & \textbf{6.17} & \textbf{2.30} & \textbf{2.300} \\
\bottomrule
\end{tabular}
`}
\caption{Results on MNIST comparing hybrid model 
to discriminative model. Arrows indicate which direction is better. 
}
\label{tab:table:mnist}
\end{table}

\begin{figure*}[h]
\centering
\subfigure[Discriminative Model ($\lambda = 0$)]{ \includegraphics[width=0.7\columnwidth]{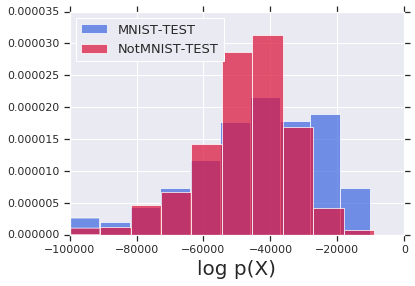}\label{discrim_gen}}
\hfil
\subfigure[Hybrid Model]{  \includegraphics[width=0.7\columnwidth]{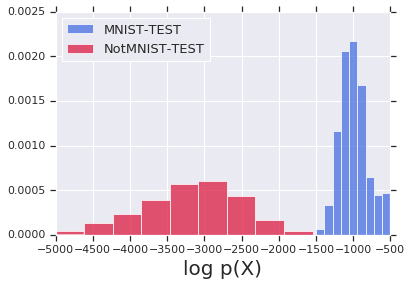}\label{hybrid_gen}}
\hfil
\subfigure[Latent Space Interpolations]{  
\includegraphics[width=0.55\columnwidth]{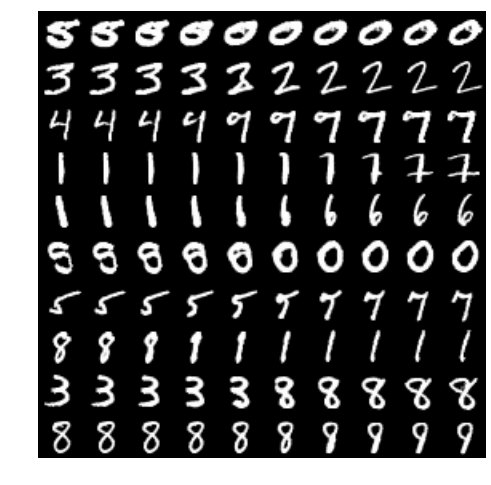}
\label{mnist_interpolations}}
\caption{Histogram of $\log p(\vx)$ on classification experiments on MNIST. The hybrid model is able to successfully distinguish between in-distribution (MNIST) and OOD (NotMNIST) test inputs. Subfigure (c) shows latent space interpolations. 
}
\label{classification_results_mnist}
\end{figure*}

Next, we compare the generative density $p(\vx)$ of the hybrid model\footnote{We report results for $\lambda=0.01/D$; higher values are qualitatively similar.} to that of the pure discriminative model ($\lambda=0$), quantifying the results in bits-per-dimension (BPD).  Since the discriminative variant was not optimized to learn $p(\vx)$, we expect it to have a high BPD for both in- and out-of distribution sets.  This experiment is then a sanity check that a discriminative objective alone is insufficient for OOD detection and a hybrid objective is necessary.  First examining the discriminative models' BPD in Table~\ref{tab:table:mnist}, we see that it assigns similar values to MNIST and NotMNIST: $81.8$ vs $87.74$ respectively.  While at first glance this difference suggests OOD detection is possible, a closer inspection of the per instance $\log p(\vx)$ histogram---which we provide in Subfigure \ref{discrim_gen}---shows that the distribution of train and test set densities are heavily overlapped.  Subfigure \ref{hybrid_gen} shows the same histograms for the DIGLM trained with a hybrid objective.  We now see conspicuous separation between the NotMNIST (red) and MNIST (blue) sets, which suggests the threshold rejection rule would work well in this case.  

Using the selective classification setup described earlier in \eqref{eq:safeprediction}, we use $p(y|\vx)$ head when $p(\vx) > \tau$ where the threshold $\tau = \min_{\vx \in X_{train}} p(\vx)$ and $p(y)$ estimated using the label counts.  The results are shown in Table~\ref{tab:table:mnist}. 
As expected, the hybrid model exhibits higher uncertainty and achieves better NLL and entropy on NotMNIST.  To demonstrate that the hybrid model learns meaningful representations, we compute convex combinations of the latent variables $\vz = \alpha \vz_1 + (1-\alpha) \vz_2$.  Figure~\ref{mnist_interpolations} shows these interpolations in the MNIST latent space.

\subsection{SVHN Classification} 
We move on to natural images, performing a similar evaluation on SVHN.  For these experiments we use a larger network of $24$ Glow blocks and employ
multi-scale factoring \citep{dinh2016density} every $8$ blocks. We use a larger Highway network containing 300 hidden units. In order to preserve the visual structure of the image, we apply only a 3 pixel random translation as data augmentation during training. The rest of the training details are the same as those used for MNIST.  We use CIFAR-10 for the OOD set.

\begin{figure*}[htbp]
\centering
\subfigure[Histogram of SVHN vs CIFAR-10 densities]{ \includegraphics[width=0.7\columnwidth]{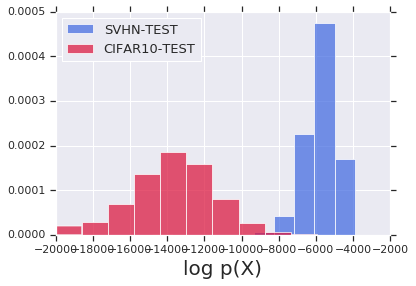}\label{classification_results_svhn}}
\subfigure[Latent Space Interpolations]{  
 \includegraphics[width=0.55\columnwidth]{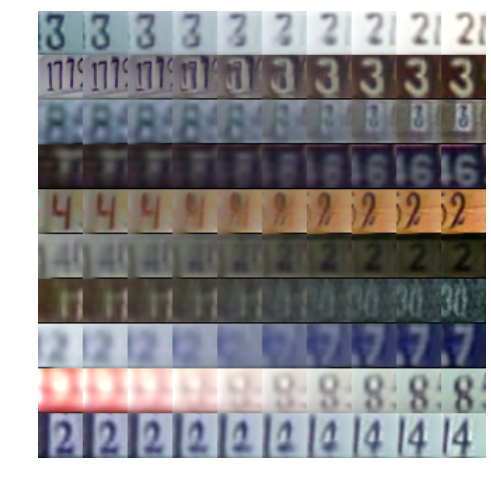} 
\label{svhn_interpolations}}
\subfigure[Confidence vs. Accuracy]{ \includegraphics[width=0.7\columnwidth]{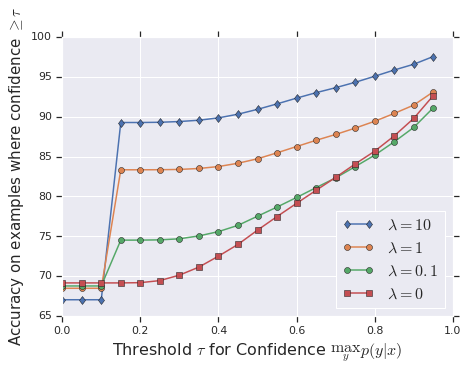}\label{confidence_accuracy_svhn}}
\caption{Subfigure (a) shows the histogram of $\log p(\vx)$ on SVHN experiments. The hybrid model is able to successfully distinguish between in-distribution (SVHN) and OOD (CIFAR-10) test inputs.  Subfigure (b) shows latent space interpolations.  Subfigure (c) shows \emph{confidence versus accuracy} plots and shows that the hybrid model is able to successfully reject OOD inputs.
}
\label{classification_results_svhn_all}
\end{figure*}

Table~\ref{tab:table:svhn} summarizes the classification results, reporting the same metrics as for MNIST. The trends are qualitatively similar to what we observe for MNIST: the $\lambda=0$ model has the best classification performance, but the hybrid model is competitive.  Figure~\ref{classification_results_svhn} reports the $\log p(\vx)$ evaluations for SVHN vs CIFAR-10. We see from the clear separation between the SVHN (blue) and CIFAR-10 (red) histograms that the hybrid model can detect the OOD CIFAR-10 samples.  Figure~\ref{svhn_interpolations} visualizes interpolations in latent space, again showing that the model learns coherent representations. Figure~\ref{confidence_accuracy_svhn} shows \emph{confidence versus accuracy} plots \citep{deepensembles},  using the selective classification rule described in Section~\ref{sec:selective:classification}, 
when tested on in-distribution and OOD, which shows that the hybrid model is able to successfully reject OOD inputs.

\begin{table}[htbp]
  \centering
    \resizebox{0.5\textwidth}{!}{
\begin{tabular}{c|c|c|c|c|c|c}
\multirow{2}{*}{Model}
&  \multicolumn{3}{ c }{\textbf{SVHN}} &  \multicolumn{3}{ |c }{\textbf{CIFAR-10}} \\ 
 &    BPD $\downarrow$ & error $\downarrow$ & NLL $\downarrow$  &    BPD $\uparrow$  & NLL $\downarrow$ & Entropy $\uparrow$ \\
\midrule
Discriminative ($\lambda=0$) &  15.40* & \textbf{4.26\%}  & \textbf{0.225} & 15.20* & 4.60 & 0.998 \\
Hybrid ($\lambda=0.1/D$) & 3.35 & 4.86\% & 0.260  & 7.06 & 5.06 & 1.153 \\
Hybrid ($\lambda=1.0/D$) & 2.40 & 5.23\% & 0.253  & 6.16 & 4.23 & 1.677 \\
Hybrid ($\lambda=10.0/D$) & \textbf{2.23} & 7.27\% & 0.268  & \textbf{7.03} & \textbf{2.69} & \textbf{2.143} \\
\bottomrule
\end{tabular}
}
\caption{Results on SVHN comparing hybrid model 
to discriminative model. Arrows indicate which direction is better. 
}
\label{tab:table:svhn}
\end{table}

\subsection{Semi-Supervised Learning}
\begin{figure}[ht]
\centering
\subfigure[Fully Supervised]{
\includegraphics[width=0.44\columnwidth]{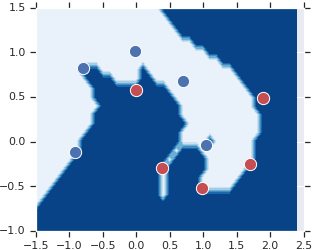}}
\subfigure[With Unlabeled Data]{
\includegraphics[width=0.44\columnwidth]{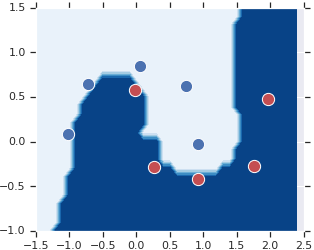}}
\subfigure{
\includegraphics[width=0.065\columnwidth]{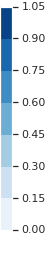}}
\caption{\textit{Half Moons Simulation}.  The decision boundary is shown for the DIGLM trained with just labeled data (left) and with unsupervised data (right).  The red and blue points are the instances that have been labeled for each class. 
}
\label{fig:halfmoon}
\end{figure}

As discussed in Section \ref{sec:ss}, one advantage of the hybrid model is the ability to leverage unlabeled data.  We first performed a sanity check on simulated data, using interleaved half moons. Figure~\ref{fig:halfmoon} shows the decision boundaries when the model is trained without unlabeled data (left) and with unlabeled data (right).  The rightmost figure shows a noticeably smoother boundary that better respects the half moon shape. 

Next we present results on MNIST when training with only $1000$ labeled points ($2\%$ of the data set) and using the rest as unlabeled data.  For the unlabeled points, we maximize $\log p(\vx)$ in the usual way and minimize the entropy for the $p(y|\vx)$ head, corresponding to \emph{entropy minimization} \citep{grandvalet2005semi}.  We also use virtual adversarial training (VAT) \citep{miyato2018virtual}, which we found to boost performance.  We chose weights on the generative model and on the VAT objective by performing grid sweeps on a validation set, see 
Appendix~\ref{sec:additional:results} 
for details.   Table~\ref{tab:table:mnist:semisup} shows the results.  We see that incorporating the unlabeled data results in an improvement from $6.61\%$ error to $0.99\%$ error, which is competitive with other SOTA approaches such as ladder networks \citep{rasmus2015semi} ($0.84\%$) and GANs \citep{springenberg2015unsupervised} ($1.73\%$).

\begin{table}[htbp]
  \centering
  \resizebox{\columnwidth}{!}{
\begin{tabular}[t]{lcc}
\toprule
Model &   MNIST-error $\downarrow$ & MNIST-NLL $\downarrow$  \\
\midrule
1000 labels only &  6.61\%  &  0.276  \\
1000 labels + unlabeled &  0.99\% &   0.069   \\
All labeled & \textbf{0.73}\% & \textbf{0.035}   \\
\bottomrule
\end{tabular}
}
\caption{Results of hybrid model for semi-supervised learning on MNIST. Arrows indicate which direction is better.
}
\label{tab:table:mnist:semisup}
\end{table}

\section{Discussion}
We have presented a neural hybrid model created by combining deep invertible features and GLMs.  We have shown that this model is competitive with discriminative models in terms of predictive performance but more robust to out-of-distribution inputs and non-stationary problems.  The availability of exact $p(\vx,y)$ allows us to simulate additional data, as well as compute many quantities readily, 
which could be useful for downstream applications of generative models, including but not limited to semi-supervised learning, active learning, and domain adaptation.

There are several interesting avenues for future work.  Firstly, recent work has shown that 
deep generative models can assign higher likelihood to OOD inputs \citep{arxivVersion, choi2018generative}, meaning that our rejection rule is not guaranteed to work in all settings. This is a challenge not just for our method but for all deep hybrid models. 
The DIGLM's abilities may also be improved by considering flows constructed in other ways than stacking ACLs.  Recently proposed continuous-time flows \citep{grathwohl2018ffjord} and invertible residual networks \citep{behrmann2018invertible} may prove to be more powerful that the Glow transform that we use, thereby improving our results.
  Lastly, we have only considered KL-divergence-based training in this paper. Alternative training criteria such as 
  Wasserstein distance could potentially further improve performance. 

\subsubsection*{Acknowledgements}
We thank
Danilo Rezende
for helpful feedback.

\bibliographystyle{icml2019}
\bibliography{references}

\clearpage
\newpage
{\textbf{\Large{\centering Supplementary Materials}}}
\appendix
\setcounter{figure}{0}
\setcounter{table}{0}
\makeatletter 
\renewcommand{\thefigure}{S\@arabic\c@figure}
\renewcommand{\thetable}{S\@arabic\c@table}
\makeatother

\section{Background on Affine Coupling Layers}\label{sec:acl}
One ACL performs the following operations \citep{dinh2016density}: \begin{enumerate}
    \item \textit{Splitting}: $\vx$ split it at dimension $d$ into two separate vectors $\vx_{:d}$ and $\vx_{d:}$ (using Python list syntax).  
    \item \textit{Identity and Affine Transformations}: Given the split $\{ \vx_{:d}, \vx_{d:} \}$, each undergoes a separate operation:  
    \begin{align}
        &\text{\texttt{identity}}: \vh_{:d} = \vx_{:d}\\ &\text{\texttt{affine}}: \vh_{d:} = t(\vx_{:d}; \vphi_{t}) + \vx_{d:} \odot \exp\{ s(\vx_{:d}; \vphi_{s}) \}  \nonumber
    \end{align} 
    where $t(\cdot)$ and $s(\cdot)$ are translation and scaling operations with no restrictions on their functional form.  We can compute them with neural networks that take as input $\vx_{:d}$, the other half of the original vector, and since $\vx_{:d}$ has been copied forward by the first operation, no information is lost that would jeopardize invertibility.
    \item \textit{Permutation}: Lastly, the new representation $\vh = \{ \mathbf{h}_{:d}, \mathbf{h}_{d:} \}$ is ready to be either treated as output or fed into another ACL.  If the latter, then the elements should be modified so that $\mathbf{h}_{:d}$ is not again copied but rather subject to the affine transformation.  \citet{dinh2016density} simply exchange the components (i.e. $\{ \mathbf{h}_{d:}, \mathbf{h}_{:d} \})$) whereas \citet{kingma2018glow} apply a $1\times1$ convolution, which can be thought of as a continuous generalization of a permutation.   
\end{enumerate}  Several ACLs are composed to create the final form of $f(\vx; \vphi)$, which is called a \textit{normalizing flow} \citep{rezende2015variational}.  Crucially, the Jacobian of these operations is efficient to compute, simplifying to the sum of all the scale transformations: 
\begin{equation*}\begin{split}
 \log \left| \frac{\partial \vf_{\vphi}}{ \partial \vx} \right| &= \log \exp \left \{ \sum_{l=1}^{L} s_{l}(\vx_{:d}; \vphi_{s,l}) \right \} \\ &= \sum_{l=1}^{L} s_{l}(\vx_{:d}; \vphi_{s,l})     
\end{split}\end{equation*}
where $l$ is an index over ACLs.  The Jacobian of a $1\times 1$ convolution does not have as simple of an expression, but \citet{kingma2018glow} describe ways to reduce computation.  
 
 \section{Additional Semi-Supervised Results} \label{sec:additional:results}
\subsection{Hyperparameters for Semi-supervised learning}
The hyperparameters are described in Table~\ref{tab:table:hyperparam:semisup}. 
 \begin{table}[htbp]
  \centering
  \resizebox{\columnwidth}{!}{
\begin{tabular}[t]{lc}
\toprule
Hyper-parameter &   Grid values  \\
\midrule
 Dropout rate & 0, 0.2, 0.5 \\
${\epsilon}_{\textsf{VAT}}$  for Virtual Adversarial Training & 1, 5 \\
 $\lambda_{\textsf{EM}}$ for Entropy Minimization loss &  0, 0.3, 1, 3 \\
 $\lambda_{\textsf{VAT}}$ for Virtual Adversarial Training loss & 0, 0.3, 1, 3 \\
\bottomrule
\end{tabular}
}
\caption{Hyperparameters for the semi-supervised learning experiments.}
\label{tab:table:hyperparam:semisup}
\end{table}

\subsection{Semi-supervised learning on SVHN}

The results are shown in Table~\ref{tab:table:svhn:semisup}. Similar to the semi-supervised  results on MNIST,  
we observe that our model can effectively leverage unlabeled data. 
 \begin{table}[htbp]
  \centering
  \resizebox{\columnwidth}{!}{
\begin{tabular}[t]{lcc}
\toprule
Model &   SVHN-Error $\downarrow$ & SVHN-NLL $\downarrow$  \\
\midrule
1000 labels only &   19.26\%  &    0.78 \\
1000 labels + unlabeled &   5.90\% &    0.38   \\
All labeled & 4.86\% & 0.26  \\
All labeled + unlabeled & \textbf{2.80}\% & \textbf{0.17 }   \\
\bottomrule
\end{tabular}
}
\caption{Results of hybrid model for semi-supervised learning on SVHN. Arrows indicate which direction is better. 
}
\label{tab:table:svhn:semisup}
\end{table}

\section{Extensions} \label{sec:extensions}

\subsection{Mixed Effects Model}
To model heteroscedastic noise, we can also add ``random effects'' to the model at the latent level.  The model is then:  \begin{align*}
   \mathbb{E}[y_{n} | \mathbf{x}_{n}] &= g^{-1}\left(\boldsymbol{\beta}^{T}f(\mathbf{x}_{n}; \boldsymbol{\phi}) + \mathbf{u}^{T}\mathbf{a}_{n} \right), \nonumber\\ \mathbf{u} &\sim p(\mathbf{u}), \ \ \ \mathbf{a}_{n} \sim p(\mathbf{a})
\end{align*} where $\mathbf{a}_{n}$ is a vector of random effects associated with the fixed effects $\mathbf{x}_{n}$, and $\mathbf{u}$ are the corresponding parameters in the GLM.  
Note that \citet{depeweg2018} also use random effects model to handle heteroscedastic noise, but add them at the input level.

\end{document}